\documentclass[11pt,a4paper]{article}
\usepackage[hyperref]{emnlp2020}
\usepackage{times}
\usepackage{latexsym}

\usepackage{microtype}
\usepackage{graphicx}
\usepackage{tabularx}
\usepackage{multirow}
\usepackage{multicol}
\usepackage{bm}
\usepackage{amsmath,amsfonts,amssymb}
\usepackage{paralist}

\aclfinalcopy 

\title{Reliable Evaluations for Natural Language Inference \\
based on a Unified Cross-dataset Benchmark}

\author{
  Guanhua Zhang$^{1,2}$, Bing Bai$^{1}$, Jian Liang$^{1}$, Kun Bai$^1$, Conghui Zhu$^2$, Tiejun Zhao$^2$\\
  $^1$Cloud and Smart Industries Group, Tencent, China\\
  $^2$Harbin Institute of Technology, China\\
  \texttt{\{guanhzhang,icebai,joshualiang,kunbai\}@tencent.com},\\
  \texttt{\{chzhu,tjzhao\}@hit-mtlab.net}
}

\date{}

\begin{document}
\maketitle
\begin{abstract}
Recent studies show that crowd-sourced Natural Language Inference~(NLI) datasets may suffer from significant biases like annotation artifacts. Models utilizing these superficial clues gain mirage advantages on the in-domain testing set, which makes the evaluation results over-estimated. The lack of trustworthy evaluation settings and benchmarks stalls the progress of NLI research. In this paper, we propose to assess a model's trustworthy generalization performance with cross-datasets evaluation.
We present a new unified cross-datasets benchmark with 14 NLI datasets, and re-evaluate 9 widely-used neural network-based NLI models as well as 5 recently proposed debiasing methods for annotation artifacts. Our proposed evaluation scheme and experimental baselines could provide a basis to inspire future reliable NLI research.
\end{abstract}

\section{Introduction}
\label{sec:intro}

Natural Language Inference~(NLI) aims to determine whether a hypothesis sentence could be inferred from a premise sentence, and the labels could be \texttt{entailment}, \texttt{neutral}, or \texttt{contradiction}. The development of large-scale datasets, \emph{e.g.}, SNLI~\cite{bowman2015large}, and MultiNLI~\cite{williams2018broad}, have greatly fertilized the research, and state-of-the-art models could achieve benchmark accuracies of over 90\% on the testing set of SNLI\footnote{https://nlp.stanford.edu/projects/snli/}.

However, recent studies have unveiled that these large-scale crowd-sourced datasets suffer from serious biases. The most significant one is the annotation artifacts~\cite{gururangan2018annotation}, \emph{i.e.}, the habits of crowd-sourcing workers when they write hypotheses leave clues for the labels. For example, negation words like \emph{no} and \emph{never} often suggest \texttt{contradiction}~\cite{gururangan2018annotation}. Models that capture this superficial pattern gain mirage advantages on the in-domain testing set and result in unreliable evaluation results.


Cross-datasets evaluation, \emph{e.g.}, training on SNLI while testing on other NLI datasets like SICK~\cite{marelli2014sick}, is an effective way to remove the impact of biases in the training data during evaluations. As the sources and preparations of datasets differ, they are less likely to suffer from the same kind of biases~\cite{zhang2019selection}.
Although different NLI datasets may be regarded as from different ``domains'', \emph{they also all belong to a same general domain --- the real world}~\cite{torralba2011unbiased}, which is often witnessed in the domain generalization problems~\cite{jiang2007instance,liang2019additive}.
Thus we argue that cross-datasets evaluation is a more reliable benchmark.


In this paper, we propose a new unified cross-dataset benchmark composed of 14 NLI datasets for models trained on SNLI in Section~\ref{sec:eva_scheme}. 
We largely eliminate the inflated accuracy scores caused by the dataset bias of SNLI, and try not to let the evaluated methods rely heavily on any specific single dataset. 
Under the proposed scheme, we evaluate 9 widely-used neural network-based NLI models~\cite{bowman2015large,conneau2017supervised,shen2018disan,talman2019sentence,parikh2016decomposable,wang2017bilateral,chen2017enhanced,gong2017natural,devlin2019bert} in Section~\ref{sec:eva_models}, and 5 existing debiasing methods for annotation artifacts~\cite{he2019unlearn,belinkov2019don,zhang2019mitigating,clark2019don} in Section~\ref{sec:eva_debiasing}.

The experimental results show that the cross-dataset generalization of NLI is still challenging. We hope our proposed evaluation scheme and experimental baselines could provide a basis to inspire the future development of reliable NLI research, which is discussed in Section~\ref{sec:direction}.

\section{Cross-dataset Evaluation Scheme}
\label{sec:eva_scheme}

In this section, we present the details of the proposed cross-dataset benchmark. We first introduce the used datasets, then we propose the evaluation scheme.

\subsection{Dataset Usage}

In out experiments, we use SNLI~\cite{bowman2015large} as the training set, then validate and evaluate models on other 14 datasets with different creation protocols. The datasets used are listed in Table~\ref{tab:dataset_usage}. Following \citet{poliak2018hypothesis}, we divide the datasets into \emph{human elicited}, \emph{human judged} and \emph{automatically recast}.

\begin{table}[t]
  \renewcommand\arraystretch{1.0}
  \centering
  \resizebox{0.482\textwidth}{!}{
  \begin{tabular}{cl}
    \hline
    \textbf{Creation Protocol} & \textbf{Dataset}\\
    \hline
    \multirow{2}{*}{Human Elicited} & $\text{MultiNLI}_\text{Matched}$~\cite{williams2018broad} \\
     & $\text{MultiNLI}_\text{Mismatched}$~\cite{williams2018broad} \\
    \hline
    \multirow{8}{*}{Human Judged} & SICK~\cite{marelli2014sick} \\
     & JOCI~\cite{zhang2017ordinal} \\
     & SciTail~\cite{khot2018scitail} \\
     & MPE~\cite{lai2017natural} \\
     & Diagnostic~\cite{wang2018glue} \\
     & Add-1~\cite{pavlick2016most} \\
     & CB~\cite{wang2019superglue, de2019commitmentbank}\\
     & RTE~\cite{wang2018glue}\\
    \hline
    \multirow{4}{*}{\shortstack{Automatically \\ Recast}} & DPR~\cite{rahman2012resolving, white-etal-2017-inference} \\
     & SPR~\cite{reisinger2015semantic, white-etal-2017-inference} \\
     & FN+~\cite{pavlick2015framenet, white-etal-2017-inference} \\
     & WNLI~\cite{wang2018glue,levesque2012winograd} \\
    \hline
  \end{tabular}
  }
  \caption{The datasets used for the proposed cross-dataset evaluation scheme.}
  \label{tab:dataset_usage}
\end{table}

Detailed introduction to the datasets is presented in Appendix~\ref{app:dataset_usage}. 

\subsection{Settings for Evaluation}

Up to 14 datasets are included in our cross dataset evaluation framework. For $\text{MultiNLI}_\text{Matched}$, we use the provided validation set as the testing set, and sample a validation set of the same size from the original training set. For $\text{MultiNLI}_\text{Mismatched}$, we split the original validation set equally for validation and testing. For $\text{CB}$, $\text{RTE}$ and $\text{WNLI}$, we concatenate the original training set and the validation set, and randomly split them into two equally. For other datasets, we randomly split the whole dataset as validation set and testing set.
Models will be trained on the training set of SNLI, hyper-parameters will be tuned based on the cross-dataset validation set, and final benchmark score will be the performance on the cross-dataset testing set.

For the final benchmark score, we collect the accuracy scores on all the testing sets, and report the average number of them. For those datasets with only two labels \texttt{entailment} and \texttt{not\_entailment}, we add up the predicted scores of \texttt{contradiction} and \texttt{neutral} as the score for \texttt{not\_entailment}. For dataset SciTail, since it only has \texttt{entailment} and \texttt{neutral}, we drop the scores of \texttt{contradiction} and normalize the rest as models' predictions. Other datasets share the same definitions for the labels.

By averaging the performance on all the testing sets, we could obtain more robust and reliable evaluation results for models. The scores will not rely heavily on the property of any specific single dataset, and thus may better reflect the models' performance in the real-world.

\section{Evaluation Results for NLI Models}
\label{sec:eva_models}

Based on the proposed cross-dataset evaluation scheme, we evaluated a wide range of NLI models. We first introduce the models, then present the results and our analyses.

\subsection{Baselines}

We chose three simple baselines, including:

\begin{compactitem}
    \item \textbf{Majority}. This method always outputs the label with the highest frequency in the validation set.
    \item \textbf{HOM}~\cite{gururangan2018annotation}. The hypothesis-only model uses \emph{fastText}~\cite{joulin2017bag} only with the hypotheses to predict the labels.
    \item \textbf{RF}~\cite{breiman2001random}. This is a simple baseline using Random Forest with 19 handcrafted features, including BLEU scores~\cite{papineni2002bleu}, word mover's distance~\cite{kusner2015word} and so on. Details are presented in Appendix~\ref{app:rf}.
\end{compactitem}

\subsection{Evaluated Models}

We evaluated several neural network-based models, which can be categorized into three kinds as,

\vspace*{0.35em} \noindent \textbf{Sentence Vector-based Models} \ 
They first map the premise and the hypothesis to vectors, then use the vectors to make final decisions. Their advantage is the ability to provide sentence vectors. We evaluated \textbf{LSTM}~\cite{bowman2015large}, \textbf{InferSent}~\cite{conneau2017supervised}, \textbf{DiSAN}~\cite{shen2018disan}, and \textbf{HBMP}~\cite{talman2019sentence}.

\vspace*{0.35em} \noindent \textbf{Interaction-based Models} \ 
They compare the premise and hypothesis at context level, and usually yield superior performance compared with sentence vector-based models. We evaluated \textbf{DecompAtt}~\cite{parikh2016decomposable}, \textbf{BiMPM}~\cite{wang2017bilateral}, \textbf{DIIN}~\cite{gong2017natural}, and \textbf{ESIM}~\cite{chen2017enhanced}.

\vspace*{0.35em} \noindent \textbf{Pretrained Models} \ 
We evaluated \textbf{BERT}-base~\cite{devlin2019bert}, which is pretrained on large-scale corpus, and fine-tuned on the training set of SNLI.


A more detailed introduction to the evaluated models is in Appendix~\ref{app:models}.

\subsection{Experiment Setup}

When the models were evaluated, all models were retrained out of the box, \emph{i.e.}, we did not tune the hyper-parameters except that the cross-dataset validation set is used for early stopping, so the performance on the testing set of SNLI may be different with what was reported in the paper.

For all models, pre-trained GloVe 840B 300D word embeddings~\cite{pennington2014glove} were used. Note that the vocabulary was built using SNLI together with the cross datasets, and the sequence length is set to 100 for all models.

\subsection{Evaluation Results for NLI Models}

\begin{table}[!t]
\renewcommand\arraystretch{1.0}
\centering
\resizebox{0.482\textwidth}{!}{
  \begin{tabular}{cc|cc|cc}
    \hline
    \multirow{2}{*}{\textbf{Model}} & \multirow{2}{*}{\textbf{Reported}} & \multicolumn{4}{c}{\textbf{Our implementation}} \\
    \cline{3-6}
    & & \textbf{SNLI} & \textbf{Cross} & $\mathbf{\Delta}$ & \textbf{$\mathbf{\Delta}$ / SNLI} \\
    \hline
    Majority & 33.8 & 32.8 & 47.1 & 14.3 & 43.6\%\\
    HOM & 67.0 & 66.9 & 45.8 & -21.1 & -31.5\%\\
    RF & -- & 54.0 & 53.2 & -0.8 & -1.5\%\\
    \hline
    LSTM & 77.6 & 72.7 & 49.0 & -23.7 & -32.6\%\\
    InferSent & 84.5 & 83.7 & 52.7 & -31.0 & -37.0\%\\
    DiSAN & 85.6 &  84.3 & 52.1 & -32.2 & -38.2\%\\
    HBMP & 86.6 & 84.2 & 54.2 & -30.0 & -35.6\%\\
    \hline
    DecompAtt & 86.3 & 83.1 & 54.2 & -28.9 & -34.8\%\\
    BiMPM & 86.9 & 84.3 & 55.2 & -29.1 & -34.5\%\\
    DIIN & 88.0 & 86.1 & 55.3 & -30.8 & -35.8\%\\
    ESIM & 88.0 & 87.5 & 56.0 & -31.5 & -36.0\%\\
    \hline
    BERT & -- & 90.6 & 62.2 & -28.4 & -31.3\%\\
    \hline
  \end{tabular}
}
\caption{Experimental results for NLI Models. ``\%'' is omitted for accuracy scores. The column ``Reported'' is the reported accuracy on the testing set of SNLI in the paper. Note that we are using the cross dataset development set for early stopping, so the accuracies of SNLI may be different from what are reported in the paper.}
\label{tab:result_model}
\end{table}

The evaluation results are presented in Table~\ref{tab:result_model}, and the more detailed results for all the cross-datasets are presented in Appendix~\ref{app:evaluation_details}.

From the table, we can get that the majority class accounts for about 47.1\% in the cross-dataset testing set, and the hypothesis-only model, \emph{i.e.}, HOM, performed slightly worse compared with Majority, while HOM outperformed Majority significantly on SNLI's testing set. This indicates that the inflated accuracy caused by SNLI's annotation artifacts is successfully reduced in our cross-dataset evaluation setting. On the other hand, models like LSTM, InferSent, and DiSAN could not even beat RF with handcrafted features under the cross-dataset evaluation, although on the given testing set of SNLI, they were far beyond RF. Such results show that these methods focus on SNLI's specific patterns which may not contribute to but rather hurt generalization. These results raise a concern that whether future accuracy-increments on SNLI will be reliable or not. As a conclusion, a cross-dataset evaluation is necessary to demonstrate the generalization performance of any NLI method.
Among all the evaluated models, only BERT could achieve more than 60\% accuracy, suggesting that the cross-dataset generalization performances of NLI models are far below satisfactory.
 
In terms of the performance drop $\Delta$, the accuracy of RF was observed very consistently on both SNLI and cross-dataset. While all the neural network-based method suffered from significant drop. Among them, pretrained model BERT did enjoy the least drop, both absolute and relative, indicating that pretraining can help the model capture the true pattern of NLI better. Besides, it can be observed that the drop of interaction-based models is slightly better, compared with the (relatively strong) sentence vector-based models.

\section{Evaluation Results for Debiasing Methods}
\label{sec:eva_debiasing}

Recently, debiasing methods~\cite{belinkov2019don,he2019unlearn,zhang2019mitigating,clark2019don} have been proposed to address the annotation artifacts in NLI. These methods can help the model avoid learning superficial patterns that happen to associate with the label on a particular dataset. In this section, we take stock of the proposed methods under the unified cross-dataset evaluation scheme to examine their effectiveness.

\subsection{Tested Methods}

We test the following methods, including
\begin{compactitem}
    \item \textbf{ADV}~\cite{belinkov2019don}. We evaluate the method~1 in the paper, which proposes to use adversarial training to remove the correlation between the label and the sentence encoding of the hypothesis, and thus discouraging models from ignoring the premise.
    \item \textbf{DRiFt}~\cite{he2019unlearn}. Focusing on the ``hard'' examples that the annotation artifacts cannot predict well, a robust model is trained to fit the residual of a hypothesis-only-model. The method is also refered as \emph{Bias Product} by \citet{clark2019don}.
    \item \textbf{Learned-Mixin} and \textbf{Learned-Mixin-H} \cite{clark2019don}. Based on \emph{DRiFT}, a confidence factor is introduced to determine how much to trust the predictions of the hypothesis-only model in \emph{Learned-Mixin}. An entropy penalty is added to the loss in \emph{Learned-Mixin+H} in order to prevent the factor from degrading to zero.
    \item \textbf{Weighting}~\cite{zhang2019mitigating}. The method manages to make models fit an artifact-neutral distribution by the instance weighting technique. The weights are generated from the predictions of the hypothesis-only model.
\end{compactitem}

\subsection{Evaluation Scheme}

We implement a one-layer 300D BiLSTM-based siamese model with max pooling as the baseline and apply all the debiasing methods on it. All hyper-parameters are chosen by the performance on the proposed cross-dataset validation set. Details are in Appendix~\ref{app:debiasing}.

\subsection{Evaluation Results}

\begin{table}[!t]
  \renewcommand\arraystretch{1.0}
  \centering
  \resizebox{0.45\textwidth}{!}{
  \begin{tabular}{c|cccc}
    \hline
    \bf Method & \bf SNLI & \bf Impr & \bf Cross & \bf Impr\\
    \hline
    BiLSTM & 79.6 & -- & 49.4 & -- \\
    +ADV & 78.4 & -1.5\% & 50.5 & +2.2\%\\
    +DRiFt & 71.3 & -10.4\% & 52.7 & +6.7\% \\
    +Learned-Mixin & 56.4 & -29.1\% & 51.1 & +3.4\% \\
    +Learned-Mixin+H & 55.0 & -30.9\% & 51.0 & +3.2\% \\
    +Weighting & 75.4 & -5.3\% & 49.9 & +1.0\%\\
    \hline
  \end{tabular}
  }
  \caption{Evaluation Results for the debiasing methods. The column ``Impr'' is the relative improvement over the baseline. ``'\%' is omitted for accuracy scores.}
  \label{tab:debiasing}
\end{table}

The evaluation results are presented in Table~\ref{tab:debiasing}. 
From the results, we can find that all debiasing methods can effectively improve models' performances on the cross dataset evaluation, while the accuracies in the original SNLI testing set drop more or less. Among the results, we find that \emph{DRiFt} brings the highest improvement in cross-dataset testing. The results conform that models are over-estimated on the the in-domain evaluation because of the dataset biases, and demonstrate that the debiasing methods can help models gain better generalization ability to the real world. 

\section{Possible Directions for Future Research}
\label{sec:direction}

Given that large-scale crowd-sourced datasets often suffer from significant biases, the following topics may be the possible directions for future research with the cross dataset evaluation.

\vspace*{0.35em} \noindent \textbf{Robust NLI Models} \ 
The differences of sentence vector-based models and interaction-based models inspire us to develop robust NLI methods. We can make the model focus more on the semantic relationship between sentences with certain neural architectures, and thus better generalization performance could be achieved without explicit modeling for dataset biases.


\vspace*{0.35em} \noindent \textbf{Debiasing NLI Methods} \ 
It is also encouraged to develop NLI methods with explicit modeling to handle dataset biases to boost generalization performance without extra training resources.

\vspace*{0.35em} \noindent \textbf{Multi-task Learning for NLI} \ 
BERT demonstrate strong generalization performance in our experiments, which inspire us to the Multi-Task Learning~(MTL)~\cite{ruder2017overview,liu2019mtdnn} for NLI.
With appropriate auxiliary tasks~(\emph{e.g.}, text coherence~\cite{nishida2018coherence}) and effective MTL methods, models may focus more on useful/generalizable patterns instead of overfit the dataset biases.

\vspace*{0.35em} \noindent \textbf{Meta-learning for NLI} \ 
\citet{finn2017model} proposed the model-agnostic meta-learning method, whose idea can be summarized as ``learning to learn.'' We may borrow the idea and perform ``learning on the biased dataset to learn on the (relatively) unbiased dataset'', thus we can utilize the large-scale human elicited datasets for training better models.


\section{Conclusion}
\label{sec:conclusion}

Dataset biases are challenging the NLI datasets and systems, not only when we are training the models, but also when the models are being evaluated. Models capturing the biases earn mirage advantages during the biased evaluations, while failing to perform well in the real-world. In this paper, we present a more trustworthy cross-dataset evaluation scheme, and re-evaluate 9 NLI models as well as 5 debiasing methods. We further discuss the possible directions for future NLI research with the proposed cross-dataset evaluation scheme. We suggest that researchers could pay more attention to cross-dataset generalization benchmarks.

\bibliography{emnlp2020}
\bibliographystyle{acl_natbib}

\clearpage

\appendix

\begin{table*}[t]
  \renewcommand\arraystretch{1.15}
  \centering
  \resizebox{0.95\textwidth}{!}{
  \begin{tabular}{c|c|c|c|c|c|c|c}
    \hline
    \textbf{Creation Protocol} & \textbf{Dataset} & \textbf{Validation Size} & \textbf{Test Size} & \textbf{Entailment} & \textbf{Not-Entailment} & \textbf{Contradiction} & \textbf{Neutral}\\
    \hline
    \multirow{2}{*}{Human Elicited} & $\text{MultiNLI}_\text{Matched}$ & 9815 & 9815 & 32.8\% & - & 32.4\% & 34.8\%\\
    & $\text{MultiNLI}_\text{Mismatched}$ & 4916 & 4916 & 35.2\% & - & 33.0\% & 31.8\%\\
    \hline
    \multirow{7}{*}{Human Judged} & SICK & 4963 & 4964 & 28.8\% & -- & 14.7\% & 56.5\%\\
     & JOCI & 2327 & 2328 & 16.3\% & -- & 26.1\% & 57.6\%\\
     & SciTail & 10000 & 17026 & 37.4\% & -- & -- & 62.6\%\\
     & MPE & 4999 & 5000 & 32.5\% & -- & 41.5\% & 26.0\%\\
     & Diagnostic & 552 & 552 & 41.7\% & -- & 23.4\% & 35.0\%\\
     & Add-1 & 1897 & 1897 & 15.7\% & 84.3\% & -- & --\\
     & CB & 153 & 153 & 45.1\% & -- & 48.0\% & 6.9\%\\
    \hline
    \multirow{5}{*}{\shortstack{Automatically \\ Recast}} & DPR & 1830 & 1831 & 50.0\% & 50.0\% & -- & --\\
     & SPR  & 10000 & 144607 & 34.8\% & 65.2\% & -- & --\\
     & FN+ & 10000 & 144604 & 43.4\% & 56.6\% & -- & --\\
     & RTE & 2876 & 2876 & 50.4\% & 49.6\% & -- & --\\
     & WNLI & 353 & 353 & 48.6\% & 51.4\% & -- & --\\
    \hline
  \end{tabular}
  }
  \caption{The datasets used for the proposed evaluation scheme.}
  \label{tab:dataset_statistic}
\end{table*}

\section{Introduction to the Used Datasets}
\label{app:dataset_usage}

Here we introduce all the used cross datasets as follows.

\begin{itemize}
    \item \textbf{MultiNLI}~\cite{williams2018broad}. MultiNLI is a human-elicted dataset. Compared with SNLI, it's with a more diverse variety of text styles and topics. Two testing set are provided, \emph{i.e.}, \emph{Matched} and \emph{Mismatched}.
    \item \textbf{SICK}~\cite{marelli2014sick}. The Sentences Involving Compositional Knowledge~(SICK) dataset is an English benchmark which include many examples of the lexical, syntactic and semantic phenomena that semantic models are expected to account for.
    \item \textbf{JOCI}~\cite{zhang2017ordinal}. JHU Ordinal Common-sense Inference~(JOCI) is a collection of diverse common-sense inference examples. We convert the labels into NLI tags following \citet{poliak2018hypothesis}.
    \item \textbf{SciTail}~\cite{khot2018scitail}. SciTail is the first NLI dataset created solely from natural sentences that already exist independently in the wild, rather than sentences authored specifically for the entailment task. The hypotheses are from science questions and the corresponding answer candidates, and the premises are from relevant web sentences retrieved from a large corpus.
    \item \textbf{MPE}~\cite{lai2017natural}. This dataset requires inference over multiple premise sentences, and trivial lexical inferences are minimized.
    \item \textbf{Add-1}~\cite{pavlick2016most}. The premise and the hypothesis in Add-1 differ only by the atomic insertion of an adjective, and only straight-forward examples are included.
    \item \textbf{DPR}~\cite{rahman2012resolving, white-etal-2017-inference}. The Definite Pronoun Resolution~(DPR) dataset targets an NLI model's ability to perform anaphora resolution.
    \item \textbf{SPR}~\cite{reisinger2015semantic, white-etal-2017-inference}. SPR is based on the seminal theory of proto-roles proposed by~\citet{dowty1991thematic}.
    \item \textbf{FN+}~\cite{pavlick2015framenet, white-etal-2017-inference}. FrameNet+~(FN+) is an expanded version of FrameNet. It contains an additional 22K lexical units.
    \item \textbf{Diagnostic} Diagnostic is designed to for analysis on a series of linguistic phenomenons by \cite{wang2018glue}. The dataset is symmetrical, \emph{i.e.}, the premise and the hypothesis in every pair is compared conversely in another.
    \item \textbf{CB} The dataset is collected from the CommitmentBank~\citep{de2019commitmentbank}, and transfered into NLI dataset by \cite{wang2019superglue}.
    \item \textbf{RTE} The dataset is converted from four annual challenges for Recognizing Textual Entailment task~\cite{dagan2005pascal, bar2006second, giampiccolo2007third, bentivogli2009fifth} and we use the same setting with \cite{wang2018glue}. 
    \item \textbf{WNLI} The dataset is gathered from the Winograd Scheme Reading Comprehension Challenge~\citep{levesque2012winograd}, in which the task is to determine the referent of the specific pronoun in each text. It is converted into a NLI dataset by replacing the ambiguous pronoun with each possible referent as hypotheses~\cite{wang2018glue}.
\end{itemize}

\begin{table*}[!t]
  \renewcommand\arraystretch{1.15}
  \resizebox{\textwidth}{!}{
  \begin{tabular}{c|c|c|c|c|c|c|c|c|c|c|c|c|c|c|c|c}
    \hline
    \bf Model & \bf MM & \bf MMis & \bf SICK & \bf JOCI & \bf SciTail & \bf DPR & \bf SPR & \bf FN+ & \bf MPE & \bf Add-1 & \bf Diagnostic & \bf CB & \bf RTE & \bf Wnli & \bf Acc Mean & \bf SNLI\\
    \hline
    \bf Majority & 31.8 & 31.8 & 56.8 & 58.4 & 62.9 & 48.4 & 65.2 & 56.6 & 26.1 & 83.8 & 34.1 & 5.9 & 49.3 & 48.7 & 47.1 & 32.8\\
    \bf HOM & 43.7 & 44.2 & 34.3 & 36.3 & 56.0 & 48.3 & 53.0 & 56.2 & 35.9 & 83.2 & 34.0 & 19.0 & 49.1 & 47.9 & 45.8 & 66.9\\
    \bf RF & 45.4 & 47.3 & 45.7 & 36.2 & 70.2 & 48.2 & 64.4 & 55.8 & 44.5 & 83.7 & 43.4 & 48.5 & 60.3 & 51.3 & 53.2 & 54.0\\
    \hline
    \bf LSTM & 45.0 & 44.9 & 42.0 & 40.7 & 60.8 & 49.7 & 61.7 & 53.1 & 46.9 & 66.9 & 39.4 & 35.5 & 50.0 & 49.4 & 49.0 & 72.7\\
    \bf InferSent & 54.4 & 55.6 & 54.0 & 40.3 & 61.9 & 50.7 & 43.8 & 50.4 & 57.9 & 70.4 & 46.6 & 44.1 & 57.7 & 50.0 & 52.7 & 83.7\\
    \bf DiSAN & 55.9 & 57.3 & 50.8 & 43.1 & 58.1 & 48.6 & 49.9 & 54.9 & 56.5 & 69.6 & 43.8 & 34.3 & 54.7 & 51.6 & 52.1 & 84.3\\
    \bf HBMP & 57.0 & 58.2 & 51.5 & 44.4 & 62.0 & 48.4 & 46.6 & 54.7 & 58.2 & 79.8 & 44.2 & 44.1 & 58.1 & 51.0 & 54.2 & 84.2\\
    \hline
    \bf DecomAtt & 58.0 & 59.6 & 50.6 & 44.8 & 62.9 & 51.1 & 42.8 & 53.1 & 57.7 & 74.1 & 47.8 & 43.8 & 60.8 & 51.6 & 54.2 & 83.1\\
    \bf BiMPM & 59.0 & 61.3 & 49.6 & 44.8 & 55.3 & 51.7 & 45.4 & 57.7 & 59.3 & 81.0 & 49.2 & 45.9 & 60.6 & 51.6 & 55.2 & 84.3\\
    \bf DIIN & 60.3 & 62.7 & 52.6 & 45.8 & 59.8 & 49.0 & 47.7 & 57.3 & 56.5 & 83.7 & 47.6 & 37.7 & 61.8 & 51.9 & 55.3 & 86.1\\
    \bf ESIM & 62.2 & 64.6 & 57.0 & 45.0 & 53.7 & 51.8 & 45.4 & 55.6 & 62.9 & 74.7 & 50.5 & 47.5 & 61.1 & 51.1 & 56.0 & 87.5\\
    \hline
    \bf BERT & 73.7 & 73.9 & 56.4 & 48.8 & 64.2 & 51.8 & 59.6 & 56.1 & 65.8 & 83.7 & 51.9 & 62.3 & 70.8 & 51.9 & 62.2 & 90.6\\
    \hline
  \end{tabular}
  }
  \caption{Detailed experimental results for NLI Models. ``\%'' is omitted. ``MM'' stands for $\text{MultiNLI}_\text{Matched}$, and ``MMis'' stands for $\text{MultiNLI}_\text{Mismatched}$.}
  \label{tab:detail}
\end{table*}

The size and the share of different labels are presented in Table~\ref{tab:dataset_statistic}.

\section{Details about the Tested Methods}

\subsection{Features used in \emph{Random Forest}}
\label{app:rf}

We list the features we used in the baseline \emph{Random Forest}. As mentioned above, we are using 19 hand-crafted features in total. The features are carefully chosen to \emph{compare} the hypothesis and the premise, including:

\begin{itemize}
    \item The BLEU score of the hypothesis to the premise, using n-gram length from 1 to 4, which are 4 features in total.
    \item The relative difference of the length between the hypothesis and the premise, in term of both words and characters, which are 2 features in total.
    \item The fuzzy string matching scores with \emph{FuzzyWuzzy}, including fuzz QRatio, fuzz WRatio, fuzz token set ratio, fuzz token sort ratio, fuzz partial ratio, fuzz partial token set ratio, fuzz partial token sort ratio, which are 7 features in total.
    \item The word mover's distance between the hypothesis and the premise, which is 1 feature.
    \item Difference type of distances between the vector representation of the hypothesis and the premise. The vector representation of a sentence is obtained by averaging the embeddings of the words. The distances include cosine distance, cityblock distance, canberra distance, euclidean distance and braycurtis distance, which are 5 features in total.
\end{itemize}

\subsection{Introduction to the Tested Models}
\label{app:models}

The introduction to the models that we tested are listed below.
\begin{itemize}
    \item \textbf{LSTM}~\cite{bowman2015large}. We evaluated the 100D LSTM encoders for sentences.
    \item \textbf{InferSent}~\cite{conneau2017supervised}. InferSent uses 4096D BiLSTM with max-pooling to encode sentences as vectors, then use the vectors to predict the NLI label.
    \item \textbf{DiSAN}~\cite{shen2018disan}. DiSAN is a 300D directional self-attention based sentence encoder.
    \item \textbf{HBMP}~\cite{talman2019sentence}. HBMP implements an iterative refinement strategy for sentence vectors with a hierarchy of 600D BiLSTM and max pooling layers.
    \item \textbf{DecompAtt}~\cite{parikh2016decomposable}. This model uses 200D attention to decompose the NLI problem into \emph{attend}, \emph{compare} and \emph{aggregate} steps.
    \item \textbf{BiMPM}~\cite{wang2017bilateral}. BiMPM combines two sentence encoders and employs a bilateral multi-perspective matching mechanism.
    \item \textbf{ESIM}~\cite{chen2017enhanced}. ESIM is a carefully designing sequential inference models based on chain LSTM.
    \item \textbf{DIIN}~\cite{gong2017natural}.DIIN is designed to hierarchically extracting semantic features from interaction space with DenseNet~\cite{huang2017densely}.
    \item \textbf{BERT}~\cite{devlin2019bert}. BERT is a pretrained bidirectional encoder with transformers, and yield strong performance for many tasks. We use Bert-base in our experiments.
\end{itemize}

\section{Detailed Evaluation Result}
\label{app:evaluation_details}

The detailed evaluation results are presented in Table~\ref{tab:detail}. We report the accuracy scores of all the models on different cross-datasets.

\section{Detailed Evaluation Scheme for Debiasing Methods}
\label{app:debiasing}

We use grid searching to find the best hyper-parameters for debias methods.
For \emph{ADV} method, we search the hyper-parameters $\alpha$ and $\beta$ over $\{0.1, 0.2, 0.5, 1, 2\}$.
For \emph{Weighting} method, we sweep the the smooth term  $\text{smooth}$ for weight generatiing over $\{0.1, 0.01, 0.02, 0.001, 0\}$.
For \emph{Learned Mixin+H} method, we search the entropy coefficient $w$ over $\{1, 0.5, 0.1, 0.05, 0.01, 0.005, 0.001\}$

\end{document}